# Causal Machine Learning for Patient-Level Intraoperative Opioid Dose Prediction from Electronic Health Records

Jonas Valbjørn Andersen[a,1], Anders Peder Højer Karlsen[b], Markus Harboe Olsen[c,d] and Nikolaj Krebs Pedersen[e]

[a] Department of Management, Aarhus University
[b] Department of Anaesthesia and Intensive Care, Bispebjerg Hospital
[c] Copenhagen Trial Unit, Copenhagen University Hospital – Rigshospitalet
[d] Department of Neuroanaesthesiology, Neuroscience Centre, Copenhagen University Hospital – Rigshospitalet
[e] IT University of Copenhagen

ORCiD ID: Jonas Valbjørn Andersen https://orcid.org/0000-0002-3366-5573

*Abstract*— This paper introduces the OPIAID algorithm, a novel approach for predicting and recommending personalized opioid dosages for individual patients. The algorithm optimizes pain management while minimizing opioid related adverse events (ORADE) by employing machine learning models trained on observational electronic health records (EHR) data. It leverages a causal machine learning approach to understand the relationship between opioid dose, case specific patient and intraoperative characteristics, and pain versus ORADE outcomes. The OPIAID algorithm considers patient-specific characteristics and the influence of different opiates, enabling personalized dose recommendations. This paper outlines the algorithm's methodology and architecture, and discusses key assumptions, and approaches to evaluating its performance.

## I. INTRODUCTION

Causal machine learning techniques [1] can be applied to a wide range of healthcare-related prediction tasks ranging from average to individualised effects and for both binary and continuous treatments [2]. Individualised opioid dose prediction from electronic health record (EHR) data represents a particularly challenging task as conditional average dose-response (CADR) predictions involve a range of unobserved effects conditional on both opioid dosage and individual patient-specific case characteristics [1]. While dose-response prediction is emerging in various domains within healthcare informatics [2, 3-4], CADR prediction is still relatively under-utilized.

In this paper, we propose a causal machine learning framework for CADR predictions of patient-level opioid dosage in intraoperative anaesthesia.

Opioids are central in perioperative analgesia. Existing perioperative opioid administration guidelines rely on cohort-level evidence from clinical trials. Thus, the multitude of patient-specific factors affecting individual opioid responses remain unaccounted for. Of globally 300 million surgeries annually, 30-60% of patients experience pain or opioid-related adverse events (ORADEs) 0-24 hours after surgery [5]. This fundamental healthcare challenge has severe individual and societal implications.

While opioid underdosing hinders pain relief and overdosing induces ORADEs, optimal opioid doses are considered those where pain and ORADEs are at the lowest possible combined level [6]. As such, the goal is to develop an algorithm that can estimate the response of an individual patient to a theoretically optimized opioid dose, even though this dose was not administered and its outcomes observed.

To address this complex machine learning task, the proposed algorithm involves counterfactual outcome prediction, a causal inference technique that aims to predict outcomes for hypothetical scenarios, i.e., optimal dosages conditional to individual case characteristics [1]. Specifically, we implement a utility function to minimize both pain and ORADES based on machine learning predictions of all patient-specific combinations of opioid dosage and individual case characteristics.

This complex model introduces several challenges including ensuring model generalisation across different patient populations and surgical procedures, validating the counterfactual predictions, and balancing model complexity with interpretability for clinical use. As traditional accuracy metrics like AUC, ACC, and F1 scores, are insufficient for model evaluation, we outline a method to develop custom evaluation metrics that account for the balance between pain management and ORADE prevention. Additionally, as this model could potentially influence clinical decisions and patient care guidelines, ensuring fairness, transparency, and robustness in the algorithm's predictions is crucial [7].

---

[1] Corresponding Author: Jonas Valbjørn Andersen, jva@mgmt.au.dk.

## II. CAUSAL MACHINE LEARNING METHOD

To predict the patient-level efficacy $\mu$ of a range of doses $D(t)$ of opiate $t \in T$ conditional on patient-specific individual case characteristics $X$ we leverage a conditional average dose response (CADR) model [8] following a causal machine learning approach [3]. Conditional average dose-response models are a way to estimate how different doses of various opiates affect pain and ORADES, depending on specific patient-level characteristics.

As such, the causal dependencies for CADR predictions are inherently complex. Following the problem definition by Bockel-Rickermann et al. [8], we acknowledge the challenges in predicting dose responses that arise from the causal relationships between variables $T$, $D$, $X$, and $\mu$. We can explicate this challenge in a single-world intervention graph (SWIG) [9] as illustrated in Figure 1.

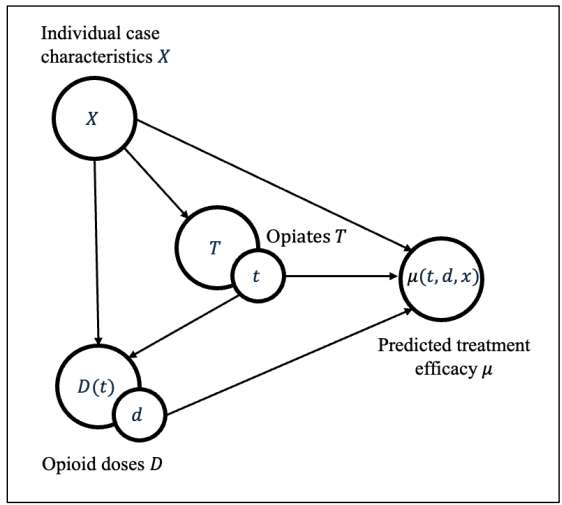

**Figure 1.** SWIG of causal dependencies in CADR prediction.

CADR predictions essentially combine predictions of two different curves for average patient responses to a range of different doses and the average efficacy of each dose conditional on individual case characteristics. A dose-response curve consists of a graph that illustrates the relationship between the dose of an opioid and the effect it has on an individual patient. For example, it might show how an opioid is more effective at higher doses up to a certain point.

Conditional average curves capture the effect of different doses accounting for individual case characteristics. In essence, it shows how the effectiveness of an opioid might change based on e.g., intraoperative procedures, age, weight, or whether the patient has certain additional health issues.

Combining dose response and conditional average curves, CADR is a graph that shows how different doses affect pain and ORADES and how those effects might vary according to individual patient characteristics. To implement this approach, we leverage a rich EHR dataset of 895.000 surgeries, including patient demographics, medical history, intraoperative details, administered opioids, and incidence and severity of pain and ORADEs.

## III. ALGORITHM ARCHITECTURE

Opiates $t \in T$ are represented as elements of an array of alternative treatments $T$ (e.g., different opiates). This is to enable comparison between different opiates or other treatments. However, $T$ will initially consist of a single element array of $T[morphine]$. Doses $D$ is a continuous indication of the dose administered indicated in mg morphine. Individual case characteristics are captured in the EHR dataset as an array $X$ of individual characteristics across the patient population. The scope of different individual case characteristics included will be based on clinical experience and an ongoing systematic literature review of demographic factors associated with postoperative pain, ORADEs and opioid requirements [10].

### A. Pain and ORADE Prediction

To predict the levels of pain and ORADEs, a range of different machine learning models, e.g., random forests [11-12], neural networks [13], and Bayesian models [14], are trained to predict the average effects of varying opioid doses on pain and ORADEs for patient-specific individual case characteristics and then compared on predictive performance. The prediction of pain as a function of treatment, doses, and individual case characteristics can be expressed as:

$$Y_{pain} = f_{pain}(T, D, X) + \epsilon_{pain} \quad (1)$$

Where $f_{pain}(\cdot)$ is the function learned by the machine learning model to predict pain levels based on the input features and $\epsilon_{pain}$ is an error term capturing unexplained variability in the EHR dataset. The estimation for ORADES follows a similar approach so that $Y_{orades} = f_{orades}(T, D, X) + \epsilon_{orades}$.

### B. Utility Function for Opioid Efficacy

A utility function is defined to capture the overall efficacy of an opiate based on machine learning predictions of pain relief and ORADEs, weighted by their relative importance. The goal of the OPIAID algorithm is to minimize both pain and ORADEs, but as clinical decisions might require HCPs and patients to make trade-offs between both parameters, we adopt a weighed approach to estimating opiate efficacy.

This weighted approach to estimating an opiate's efficacy highlights the importance of incorporating the complexities of clinical decision making into CADR estimation. However, defining $Y$ (the output) in a way that explicitly minimizes pain and ORADEs is challenging and requires a nuanced approach for several reasons.

First, pain and ORADEs are multifaceted. Pain and ORADEs are not simple, single-dimensional variables. They have varying intensities, durations, and types. As such, minimizing one outcome might inadvertently increase another. Second, estimations across patient populations involve a high degree of individual variation. Patients experience pain and side-effects differently based on their individual characteristics, tolerance, and prior experiences. A one-size-fits-all definition of $Y$ that minimizes these factors is unlikely to be universally applicable. Finally, clinical decision making involves trade-offs between pain relief and the potential ORADES associated with a specific opiate. A single optimal outcome might not exist, which is why it is important to consider HCPs' wider judgement as well as patients' individual preferences and priorities. Considering this weighed approach, we can define the utility function for analgesic efficacy as:

$$U(T, D, X) = -\big(w_{pain} \cdot P(T, D, X) + w_{orades} \cdot S(T, D, X)\big) \quad (2)$$

Where $U(T, D, X)$ represents the utility of treatment $T$ with dose $D$ for individual case characteristics $X$, $P(T, D, X)$ predicts the pain level under treatments $T$ with doses $D$ for individual case characteristics $X$, $S(T, D, X)$ predicts ORADE severity under treatments $T$ with doses $D$ for individual case characteristics $X$, and $w_{pain}$ and $w_{orades}$ are weights representing the relative importance of minimizing pain and adverse events, respectively.

*C. Personalized Dose Recommendation*

Given a set of patient-specific individual case characteristics and desired opioid dosage, the OPIAID algorithm identifies the dose that maximizes the utility function, based on the CADR model predictions as explained previously. Identifying this dose will serve as a personalized, patient-specific dose recommendation to be considered in clinical decision making. We can, therefore, define the expected patient-specific utility of a given analgesic treatment as:

$$\mu(t,d,x) = \mathbb{E}\left[U\left((Y_{pain}(t,d,x), Y_{orades}(t,d,x))\right) \mid X = x\right] \quad (3)$$

Where $\mu(t,d,x)$ represents the expected utility of treatment $t$ with dose $d$ for patient characteristics $x$, $\mathbb{E}[\cdot]$ denotes the expectation operator, averaging over potential outcomes, $U(\cdot)$ is the utility function defined in Eq. (2), $Y_{pain}(t,d,x)$ and $Y_{orades}(t,d,x)$ represent the predicted pain level and ORADEs, respectively under treatment $t$ with dose $d$ for patient-specific individual case characteristics $x$, and $\mid X = x$ indicates that the expectation is conditional on patient characteristics x.

## IV. PRELIMINARY VALIDATION

The OPIAID algorithm is validated by estimating the counterfactual predicted outcome 'optimal end-of-surgery intraoperative opioid dose'. As the optimal dose is not administered it must be estimated retrospectively based on observed responses. This estimation is based on the end-of-surgery intraoperative opioid dose and the 0-2h postoperative occurrence of pain, ORADEs and rescue analgesic requirements. In accordance with the utility function described above, the optimal dose will be the one that balances pain, ORADEs and rescue analgesia to a combined minimum, based on the assumption that pain and ORADEs are both harmful and unpleasant for the patient [15-16]. It is important to understand that the 'optimal end-of-surgery intraoperative opioid dose' is a counterfactual outcome that did not occur, as patients in the dataset were treated using standard of care.

*A. EHR Dataset*

To implement and evaluate the algorithm, we leverage a rich EHR dataset of 895.000 surgeries performed between 2017-2023 in the Capital Region of Denmark and Zealand. The database contains comprehensive data from the Health Platform (Epic Systems Corporation, WI, United States), including >50 variables for patient demographics, medical history, intraoperative details including surgical and anaesthesiologic factors, administered opioids, and incidence and severity of pain and ORADEs as well as related organizational outcomes (postoperative opioid consumption, morbidity, mortality and length of hospitalization).

*B. Internal Validation*

For the initial internal validation of the algorithm, we applied two alternative techniques to estimate the patient-level optimal end-of-surgery intraoperative opioid dose. These served as a point of comparison for the output of the algorithm predicted dose. The two machine learning-based and one rule-based methods that are employed to estimate the outcome.

1) Causal Machine Learning: Utilizing conditional average dose-response curves to predict outcomes [1]. We apply the above methods using eight different machine-learning models to predict pain and ORADEs (Multinomial Logistic Regression, K-Nearest Neighbor, Decision Tree, Random Forest, Extreme Gradient Boost, Artificial Neural Network, Support Vector Machine, and Naïve Bayes)[17-18] by using 80% of data for training and 15% for test. After optimization in several iterations, we use the remaining 5% as retention set for a final internal validation. The developed models will be evaluated based on accuracy and area under the curve (AUC) for classification, root mean square error (RMSE) for regression, and loss-curves for overfitting.

2) Proxy Marker Analysis: Using the length of PACU stay as an indicator of overall treatment success. We use PACU discharge readiness (time from PACU arrival to fulfillment of PACU discharge criteria) to assess length of PACU stay, to avoid overestimation due to delayed transport. To evaluate the relationship between acute pain management and mobility after surgery, we will extract all ambulation scores performed in the surgical ward on day 0-2, such as cumulated ambulation score (CAS) [19].

3) Clinician Evaluated Rule-Based Calculations: Applying standardized adjustments based on the patient's response to the intraoperative opioid dose administered by the nurse anesthetist. For example, increasing the 'optimal end-of-surgery intraoperative opioid dose' by 2 mg MEQ in cases of moderate pain 0–1 hour postoperatively.

Each of the three models will be executed and its outcomes assessed based on accuracy and area under the curve (AUC) for classification, root mean square error (RMSE) for regression, and loss-curves for overfitting. The two best performing methods will then be carried forward to external validation.

*C. External Validation*

The present external validation protocol adheres to the TRIPOD-P guideline and describes a prospective cohort study of 700 adult patients ($\geq$ 18 years) undergoing elective or acute surgery, under general anesthesia that require an end-of-surgery intraoperative opioid administration for postoperative pain management and a subsequent stay in the post-anesthesia care unit (PACU). Patients who receive preoperative neuraxial anesthesia or preoperative peripheral nerve blocks that provide complete analgesia for the surgical site (i.e. no requirement of an end-of-surgery intraoperative opioid administration) will be excluded.

## V. ASSESSED VARIABLES AND DEFINITIONS

The validation of the algorithm relies on estimations of optimal opioid doses based on predicting pain and ORADEs.

*A. Opioid doses*

Opioid doses are presented in morphine equivalence (MEQ). We define the opioid dose that is administered by the attending nurse anesthetist or anesthesiologist at the end of surgery as part of the intraoperative analgesic regimen for postoperative pain relief as *'clinician administered opioid dose'*. *'Algorithm opioid dose'* is the opioid dose that is recommended by the algorithm as the end of surgery opioid dose as part of the intraoperative analgesic regimen for postoperative pain relief. Finally, the *'calculated optimal dose'* is the dose which in retrospect evaluation is evaluated to be optimal based on the patient's response to the clinician administered opioid dose. The *calculated optimal dose* will be automatically calculated based on the same rule set used in the clinician evaluated rule-based model. Extra rules are applied specifically for fentanyl administrations to account for the fact that fentanyl is typically titrated with several administrations during surgery.

*B. Pain and ORADEs*

*Pain* at rest will measure patient-reported pain on the 11-point numeric rating scale (NRS 0-10). We will assess pain at 1) PACU arrival, 2) prior to each opioid dosing, and at 3) PACU discharge. *ORADEs* are defined as: nausea (none/mild/moderate/severe); vomiting (yes/no); sedation (Alert, Verbal, Pain, and Unresponsive scale) [20]; dizziness (yes/no); itching (yes/no; yes if the patient complains about itching or needs antihistamine rescue medication); urinary retention (yes/no; yes if the patient complains about not being able to urinate or has trouble urinating despite ultrasound-confirmed ≥200ml in the bladder); confusion (yes/no); hallucinations (yes/no); respiratory depression (yes/no; defined as yes in the occurrence of (1) naloxone use, (2) respiratory rate < 10 breaths per minute for at least 10 minutes which the PACU-nurse perceives as possibly opioid-induced, or (3) oxygen saturation < 90% for at least 10 minutes which the PACU-nurse perceives as possibly opioid-induced) [21]; and naloxone or antiemetic rescue medication (yes/no). To explore the patient-relevance of ORADEs, we will request participants who have encountered pain or ORADEs to evaluate these events' negative impact on recovery in the PACU [21]. For this evaluation, on day 1 an 11-point numeric scale will be utilized, with 0 indicating "no negative impact" and 10 representing "extreme negative impact".

## VI. Discussion

The OPIAID algorithm rests on a unique set of assumptions and challenges for evaluation that warrants further elaboration.

*A. Implicit Assumptions*

The OPIAD algorithm relies on several assumptions, particularly regarding causal consistency, the absence of hidden confounders, and sufficient data availability for each patient characteristic and treatment combination.

The causal consistency assumption states that the observed outcome $Y_i$ for a patient $i$ who received anaesthetic treatment $t_i$ and dose $d_i$ is the same as their potential outcome $Y_i(t_i, d_i)$. In simpler terms, it assumes that the observed outcome truly reflects the effect of the treatment received so that a decrease in pain and/or ORADEs are caused by the analgesic treatment.

The no hidden confounders assumption assumes that the chosen opiate $T$ and assigned dose $D$ are independent of the potential outcome $Y(t, d)$ given the individual case characteristics $X$. Essentially, it means that there are no other unobserved factors that influence both the treatment choice and the outcome. This assumption is crucial because it ensures that the observed association between treatment and outcome is truly causal and not due to hidden confounding variables.

The data overlap assumption states that for every patient with characteristics $x$, there is a non-zero probability of receiving any possible combination of intervention and dose. In other words, it ensures that there are enough observations for every treatment and dosage combination, allowing for meaningful comparisons across different treatment groups, including assumption of representational reporting of the data. This assumption is particularly important in observational studies based on EHR data where treatment assignment is not always randomized.

*B. Further Evaluation*

The algorithm's performance will be validated against real-world data, simulation studies, and sensitivity analysis. Any validation should follow the transparent reporting of a multivariable prediction model for individual prognosis or diagnosis (TRIPOD) statement [7]. Further research and validation with real-world data is crucial to confirm its effectiveness and ensure its successful implementation in clinical settings.

## VII. Conclusion

The OPIAID algorithm represents a significant advancement in opioid management. Its ability to personalize treatment based on individual patient characteristics offers great potential for optimizing pain relief while minimizing risks associated with opioid use. Future research will focus on refining and evaluating the OPIAID algorithm's accuracy, addressing clinical and ethical considerations, and integrating it into clinical workflows.